\UseRawInputEncoding
\documentclass[letterpaper]{article} 
\usepackage{aaai24}  
\usepackage{times}  
\usepackage{helvet}  
\usepackage{courier}  
\usepackage[hyphens]{url}  
\usepackage{graphicx} 
\usepackage{natbib}  
\usepackage{caption} 
\usepackage{algorithm}
\usepackage{algorithmic}
\usepackage{amsmath}

\usepackage{multicol}
\usepackage{cancel}
\usepackage{tabularx}
\usepackage{booktabs}
\usepackage{newfloat}
\usepackage{listings}
\DeclareCaptionStyle{ruled}{labelfont=normalfont,labelsep=colon,strut=off} 
\floatstyle{ruled}
\newfloat{listing}{tb}{lst}{}
\floatname{listing}{Listing}                              
\nocopyright 

\title{Chasing Consistency in Text-to-3D Generation from a Single Image}

\author {
    Yichen Ouyang\textsuperscript{\rm 1} \quad
    Wenhao Chai\textsuperscript{\rm 2} \quad
    Jiayi Ye\textsuperscript{\rm 1}\\
    Dapeng Tao\textsuperscript{\rm 4} \quad
    Yibing Zhan\textsuperscript{\rm 3}\footnotemark[1] \quad
    Gaoang Wang\textsuperscript{\rm 1}\footnotemark[1]
}

\affiliations {
    \textsuperscript{\rm 1}Zhejiang University \quad
    \textsuperscript{\rm 2}University of Washington \quad
    \textsuperscript{\rm 3}JD Explore Academy \quad
    \textsuperscript{\rm 4}Yunnan University
}

\usepackage{bibentry}
\usepackage{amssymb}

\begin{document}

\maketitle

\renewcommand{\thefootnote}{\fnsymbol{footnote}}
\footnotetext[1]{Corresponding authors.}

\begin{abstract}
Text-to-3D generation from a single-view image is a popular but challenging task in 3D vision.
Although numerous methods have been proposed, existing works still suffer from the inconsistency issues, including 1) semantic inconsistency, 2) geometric inconsistency, and 3) saturation inconsistency, resulting in distorted, overfitted, and over-saturated generations.
In light of the above issues, we present \textbf{Consist3D}, a three-stage framework Chasing for semantic-, geometric-, and saturation-\underline{\textbf{Consist}}ent Text-to-\underline{\textbf{3D}} generation from a single image, in which the first two stages aim to learn parameterized consistency tokens, and the last stage is for optimization.
Specifically, the semantic encoding stage learns a token independent of views and estimations, promoting semantic consistency and robustness.
Meanwhile, the geometric encoding stage learns another token with comprehensive geometry and reconstruction constraints under novel-view estimations, reducing overfitting and encouraging geometric consistency. 
Finally, the optimization stage benefits from the semantic and geometric tokens, allowing a low classifier-free guidance scale and therefore preventing oversaturation.
Experimental results demonstrate that Consist3D produces more consistent, faithful, and photo-realistic 3D assets compared to previous state-of-the-art methods.
Furthermore, Consist3D also allows background and object editing through text prompts.
\end{abstract}
\begin{figure}[!t]
\centering
\includegraphics[width=1\columnwidth]{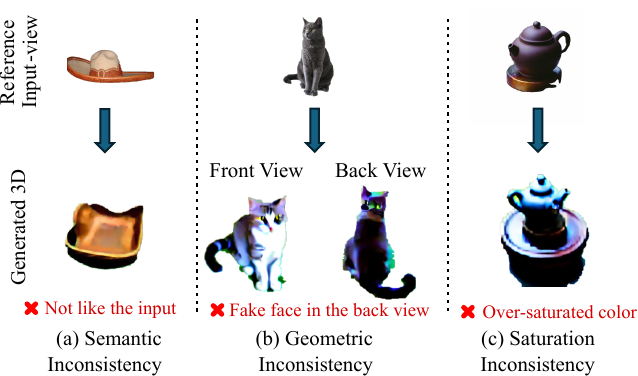}
\caption{\textbf{Inconsistency issues.} (a) The semantic inconsistency: the generated object looks like a box instead of hat.
(b) The geometric consistency: the generated cat's face exists in the back view of a cat whose face originally towards the front view.
(c) The saturation inconstancy: the rendering of the generated teapot is oversaturated compared with the original teapots color.
}
\label{inconsistency}
\end{figure}

\begin{figure*}[t]
\centering
\includegraphics[width=1\textwidth]{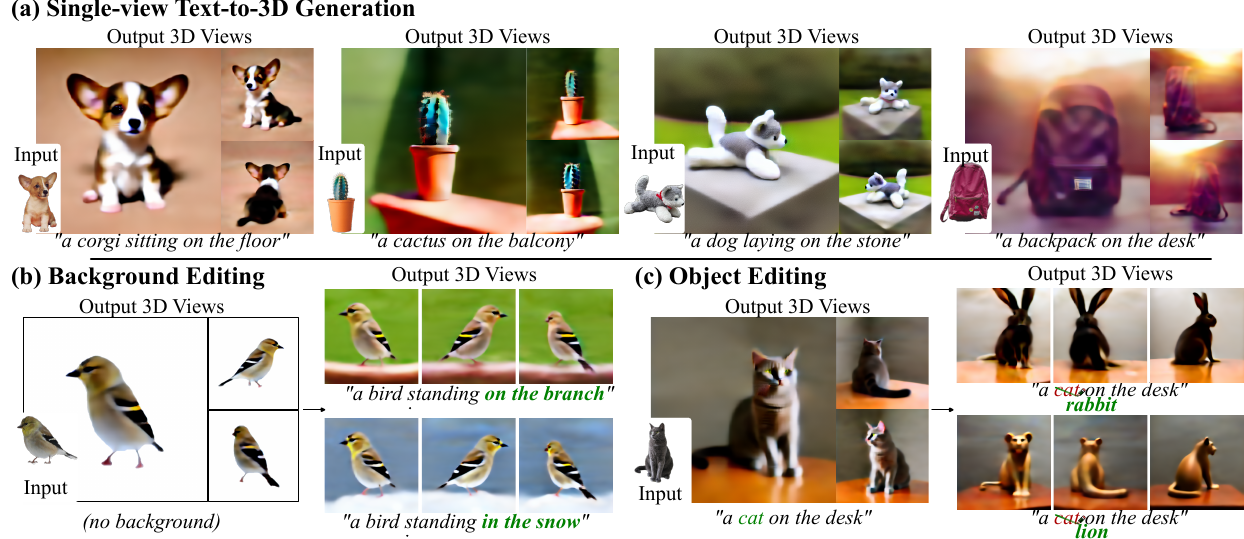}
\caption{\textbf{Effectiveness of our method.} (a) Single-view Text-to-3D generation: each case with one single input image and 3 novel views rendered from the 3D generations. (b) Background Editing: the background of the generation can be edited by prompt, and the option for no background is provided as well. (c) Object Editing: the object of the generation can be edited by prompt. For example, we can change the ``cat" into ``rabbit" or ``lion" without changing the input image.}
\label{cover}
\end{figure*}
\section{Introduction}
Recently, text-to-3D generation from a single image has emerged as an active research area, with the goal of personalizing 3D assets using a reference image. This field has been explored extensively in previous literature~\cite{raj_dreambooth3d_2023,seo_let_2023}, often relying on shape estimation or few-shot finetuning as the prior and score distillation sampling~\cite{poole_dreamfusion_2022} as the optimizer. 
Even though numerous methods have been proposed, they still suffer from inconsistency issues. For instance, 1) misguided semantics caused by inaccurate shape estimations. 2) distorted geometric caused by overfitting on the reference view. 3) over-saturated color caused by score distillation sampling.

Shape estimation methods~\cite{nichol_point-e_2022, jun_shap-e_2023, yu_points--3d_2023, sanghi_sketch--shape_2023}, including point cloud estimation, sketch estimation, etc., aim to aid text-to-3D generation by providing an estimated 3D prior for each novel view.
However, they often inaccurately estimate the 3D priors, which results in misguided semantics, especially when a single-view image is the only input, because, in such a situation, there is no enough information provided for estimating 3D priors. 
Few-shot fine-tuning methods~\cite{ruiz_dreambooth_2022, zhang2023inversion} aim at personalizing a text-to-image generative model with several images with a common subject. However, when only one single input image is provided for training, they often lead to geometric inconsistency across novel views because of overfitting on the reference single view.
The score distillation sampling method~\cite{poole_dreamfusion_2022} aims to lift the 2D generations to 3D assets. This 
lifting process needs to ensure that the generation under each viewing angle is stable enough, so the fidelity of saturation is sacrificed for stability with a high classifier-free guidance scale applied.
Therefore, current methods for generating 3D assets from a single image face challenges with inconsistency in semantics, geometry, and saturation, often resulting in distorted and over-saturated generations, as illustrated in Fig.~\ref{inconsistency}. Enhancing semantic and geometric consistency across seen and unseen views while being robust to inaccurate shape estimations, and mitigating color distortion in optimization, is imperative for achieving satisfactory 3D generation results. 

In this paper, we present \textbf{Consist3D}, a semantic-geometric-saturation \underline{\textbf{Consist}}ent approach for photo-realistic and faithful text-to-\underline{\textbf{3D}} generation from a single image. We address the three inconsistency issues (as shown in Fig.~\ref{inconsistency}) by introducing a three-stage framework, including a semantic encoding stage, a geometric encoding stage, and an optimization stage. In the first stage, a parameterized identity token is trained independently of shape priors, enhancing robustness to misguidance and relieving the semantic-inconsistency problem. In the second stage, a geometric token is trained with comprehensive geometry and reconstruction constraints, overcoming single-view overfitting issues and further enhancing geometric consistency between different views. In the third stage, the optimization process benefits from the semantic token and geometric token, allowing low classifier-free guidance (CFG) scales, therefore addressing the saturation-inconsistency issue and 
enabling background and object editing through text prompt.

The experiments highlight the strengths of Consist3D in generating high-fidelity 3D assets with robust consistency, while remaining faithful to the input single image and text prompts.
As shown in Fig.~\ref{cover} (a),  compared to baseline methods, our generated results exhibit improved consistency and more reasonable saturation. Notably, our approach enables background editing (Fig.~\ref{cover} (b)) and object editing (Fig.~\ref{cover} (c)) through text prompt, without changing the input image.
We summarize our contribution as follows:
\begin{itemize}
    \item To our knowledge, we are the first to explore the semantic-geometric-saturation consistency problems in text-to-3D generation, and accordingly, we propose \textbf{Consist3D}, an approach for consistent text-to-3D generation, background and object editing from a single image.
    \item Our Consist3D consists of a three-stage framework, including a semantic encoding stage, a geometric encoding stage, and an optimization stage, and can generate robust, non-overfitted, natural-saturated 3D results under a low classifier-free guidance scale.
    \item Extensive experiments are conducted. Compared with prior arts, the experimental results demonstrate that Consist3D produces faithful and photo-realistic 3D assets with significantly better consistency and fidelity.
\end{itemize}

\begin{figure*}[t!]
\centering
\includegraphics[width=\textwidth]{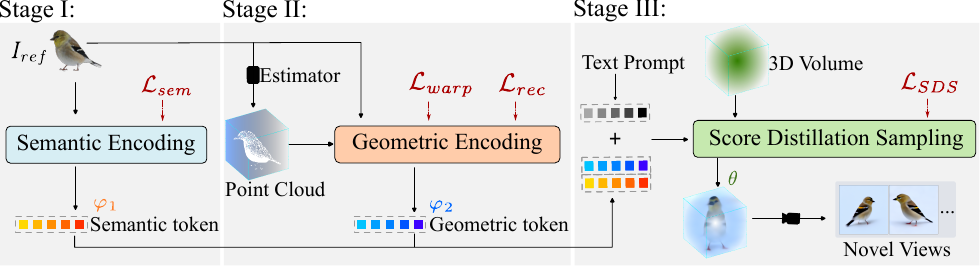} 
\caption{\textbf{Pipeline.} Stage I. A single-view image is input to the semantic encoding module, and a semantic token is trained with sem loss. Stage II. The single-view image is the input and used to estimate a point cloud as the shape guidance to apply condition on the geometric encoding module, and a geometric token is trained with warp loss and rec loss. Stage III. A randomly initialized 3D volume is the input and the two tokens trained previously is utilized together with tokenized text prompt as the condition, and this 3D volume is trained into a 3D model faithful to the reference single image.} 
\label{pipeline}
\end{figure*}
\section{Related Works}
\subsection{Personalized Text-to-Image Generation}
Text-to-image (T2I) generative models~\cite{rombach_high-resolution_2022,podell_sdxl_2023,saharia_photorealistic_2022,ramesh_hierarchical_2022,xue_raphael_2023} have significantly expanded the ways we can create 2D images with text prompt in multi-modality field~\cite{chai2022deep}. With text-to-image (T2I) synthesis enhanced by controllable denoising diffusion models~\cite{cao2023image, huang2023composer, kawar2023imagic,huang2023collaborative}, personalizing text-to-image generation has become the emerging focus of research, which aims to generate images faithful to a specific subject. 
This area has seen considerable exploration~\cite{dhariwal_diffusion_2021,ho_classifier-free_2022, zhang_adding_2023, gal_image_2022,ruiz_dreambooth_2022, wu_tune--video_2023}.
For example, textual inversion methods~\cite{yang2023controllable,zhang2023inversion,huang2023reversion,voynov2023p+} learn parameterized textual descriptions from a set of images sharing a common subject. This extends the T2I tasks to image-to-image generation, essentially realizing subject-driven personalization. 
To enhance textual inversions efficiently, there currently emerges a lot of few-shot finetuning approaches. 
Typically, DreamBooth~\cite{ruiz_dreambooth_2022} learns parameterized adapters (\textit{i.e.}, LoRA~\cite{hu_lora_2021}) for the generative network, instead of parameterizing the textual descriptions.
In another direction, ControlNet~\cite{zhang_adding_2023}, Composer~\cite{huang2023composer}, and T2I-Adapter~\cite{mou2023t2i} offer guidance to diffusion models, facilitating custom-defined constraints over the generation process, and yielding controllable personalization.

\subsection{Personalized Text-to-3D Generation}
Personalized text-to-3D generation has gained interest by extending successful personalized T2I models, aiming at generating 3D assets from a few images~\cite{raj_dreambooth3d_2023, xu_dream3d_2023}. Most current approaches~\cite{metzer_latent-nerf_2022,raj_dreambooth3d_2023} apply few-shot tuning (e.g., DreamBooth) for personalization and score distillation sampling (SDS)~\cite{poole_dreamfusion_2022} for optimization.
A generalized DreamFusion approach combines few-shot tuning on a few images for personalization and estimations from Zero-1-to-3~\cite{liu_zero-1--3_2023} as the shape priors, followed by SDS optimization. However, shape priors estimated by Zero-1-to-3 are often view-inconsistent, resulting in low-quality generations.
Another work, DreamBooth3D~\cite{raj_dreambooth3d_2023} enables personalized 3D generation from 3-5 images via joint few-shot tuning and SDS optimization. However, when the input views are decreased to 1, overfitting on limited views leads to reconstruction failures and geometric inconsistency for novel views.
Generating personalized 3D assets from only one single input image remains challenging~\cite{cai_diffdreamer_2023, gu_learning_2023, deng_nerdi_2023, gu_nerfdiff_2023, xing_semi-supervised_2022, lin_vision_2022}. 
3DFuse enables one-shot tuning on a single image and an estimated point cloud as guidance for a ControlNet for personalization, which performs together with SDS optimization. However, semantic and geometric inconsistency across views persists, as the point cloud estimation lacks accuracy, and one-shot tuning overfits the given view. This results in blurred, low-fidelity outputs.
Score distillation sampling (SDS) optimizes 3D volume representations using pretrained text-to-image models, first introduced in DreamFusion~\cite{poole_dreamfusion_2022}. With the introduction of SDS, high-quality text-to-3D generation has been achieved in many previous works~\cite{lin_magic3d_2023,tang_make-it-3d_2023,tsalicoglou_textmesh_2023, chen_fantasia3d_2023, wang_prolificdreamer_2023}. The insight of SDS is that under high classifier-guidance (CFG) scale, the generation of T2I model is stable enough under each text prompt, therefore enabling the 3D volume to converge. However, current works find that high CFG scale harms quality of the generations, leading to over-saturated results~\cite{wang_prolificdreamer_2023}.

\section{Method}
\subsection{Overview}
The input to our approach is a single image $I_{ref}$ and a text prompt $y$. We aim to generate a $\theta$ parameterized 3D asset that captures the subject of the given image while being faithful to the text prompt. 
To achieve consistent 3D generation in the encoding process, we learn semantic consistency token and geometric consistency token, parameterized by $\varphi_1$ and $\varphi_2$, respectively.
Overall, the parameters we need to optimize are $\varphi_1,\varphi_2,\theta$, and the optimization goal can be formulated as follows, 
\begin{equation}
\min_{\varphi_1,\varphi_2,\theta}\mathcal{L}(g(\theta, c),\epsilon(I_{ref}, y, y_{\varphi_1,\varphi_2}, c)),
\label{overall}
\end{equation}
where $\mathcal{L}$ is the loss function, $c$ is the camera view, $g$ is a differential renderer, and $\epsilon$ is the diffusion model used to generate image using both text prompt $y$ and learned prompt $y_{\varphi_1,\varphi_2}$ under the given view.

To facilitate the optimization of the parameters $\varphi_1,\varphi_2,\theta$, we adopt two encoding stages and one score distillation sampling stage in our pipeline (Fig. \ref{pipeline}). 
In the first stage, we propose semantic encoding and fine-tune a pretrained diffusion model $\epsilon_{pretrain}$ to learn a semantic token parameterized by $\varphi_1$, aiming at encapsulating the subject of the given image.
In the second encoding stage, we propose geometric encoding to learn a geometric token parameterized by $\varphi_2$, with carefully designed geometry constraints and reconstruction constraints. 
In the score distillation sampling stage, we propose a low-scale optimization for $\theta$ parameterized 3D volume presentations, benefited specifically from the enhanced consistency with the proposed tokens.

\subsection{Semantic Encoding}

The semantic encoding stage aims to learn the semantic token parameterized by $\varphi_{1}$. The semantic token can be further incorporated with the text prompt to faithfully reconstruct the reference view image $I_{ref}$ with consistent semantics.
Specifically, we use the single image $I_{ref}$ as the input to do one-shot fine-tuning to obtain the semantic token parameterized by $\varphi_{1}$ to represent the given image as follows, 
\begin{equation}
\begin{split}
\min_{\varphi_1}\mathcal{L}_{sem}&(\varphi_1):=\\
\mathbb{E}_{x,\epsilon ,t}[&w(t)\cdot\lVert{\epsilon_{pretrain}(I_{ref},t,y_{\varphi_1})-\epsilon_t}\rVert_2^2],
\label{fml0}
\end{split}
\end{equation}
where 
$y_{\varphi_1}$ is a prompt containing the semantic token, $\epsilon_{pretrain}$ represents the pretrained stable diffusion model, $\epsilon_t$ is the noise scheduled at time step $t$, and $w(t)$ is the scaling factor which will be discussed in detail in Section~\ref{sec:stage_3}.

We use the same training setting to DreamBooth~\cite{ruiz_dreambooth_2022}, which enables few-shot personalization of text-to-image models using multiple reference images of a subject. Specifically, we adopt DreamBooth for one-shot personalization, which optimizes $\varphi_1$ by Eq.{~\ref{fml0}} to identify the single-view image. 
Notably, with only one image $I_{ref}$ as the input, naive DreamBooth tends to overfit not only the subject but also the view of the reference image, leading to inconsistent generations under novel views. To address this, we propose the second encoding stage to improve the geometric consistency.

\begin{figure}[t!]
\centering
\includegraphics[width=\columnwidth]{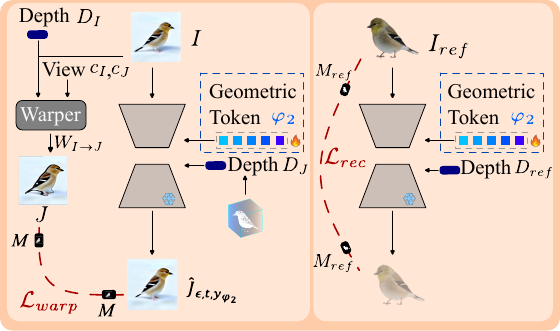} 
\caption{\textbf{Geometric encoding.} We adopt ControlNet with depth guidance for the generation. The training object is $\mathcal{L}_{warp}$ and $\mathcal{L}_{rec}$. The $\mathcal{L}_{warp}$ calculated loss between two neighboring views with warp mask under novel views, and the $\mathcal{L}_{rec}$ calculated loss between the single input image and the generation with reference mask under reference view.} 
\label{geometric}
\end{figure}

\subsection{Geometric Encoding}
In the second stage, we propose geometric encoding (Fig.~\ref{geometric}), which aims to solve the overfitting and inconsistency issues by encapsulating warp and reconstruction consistency into what we term geometric token, parameterized by $\varphi_2$. 

To achieve warp and semantic consistency, the overall objective $\mathcal{L}_{geometric}$ combines the two terms $\mathcal{L}_{warp}$ and $\mathcal{L}_{rec}$ (Eq.{~\ref{fml7}}). Notably, the consistency token from this encoding stage does not contain standalone semantics. Due to depth guidance, its semantic is conditioned on view $c$, encapsulating inherent 3D consistency of generation. By incorporating this token into prompts, we enhance geometric consistency of diffusion model outputs across different views.
\begin{equation}
\mathcal{L}_{geometric}(\varphi_2 \mid c_I)=\mathcal{L}_{warp}(\varphi_2 \mid c_I)+ \mathcal{L}_{rec}(\varphi_2 \mid c_{ref}),
\label{fml7}
\end{equation}
where $c_I$ defines the sampled camera view for image $I$, $c_{ref}$ is the given input reference view. The warp loss and the reconstruction loss are demonstrated as follows.

\subsubsection{Warp Loss}
The warp loss aims to ensure a consistent transition between two camera views, $c_I$ and $c_J$, with a learnable geometric token parameterized by $\varphi_2$. The loss is formulated as follows,
\begin{equation}
\begin{split}
\min_{\varphi_2}\mathcal{L}_{warp}&(\varphi_2 \mid c_I):=\mathbb{E}_{x,\epsilon ,t,c_I}[w(t)\cdot\\
&\lVert{(\hat{J}_{\epsilon,t,y_{\varphi_2}} - \mathcal{W}_{I\to{J}}(I, D))\cdot{M}}\rVert_2^2],
\end{split}
\label{fml4}
\end{equation}
where $\hat{J}_{\epsilon,t,y_{\varphi_2}}$ is the generated image from the diffusion model under the view $c_J$ guided by the learnable geometric token $\varphi_2$, $\mathcal{W}_{I\to{J}}(I, D)$ is the warp operator that transfers the image $I$ from the view $c_I$ to the view $c_J$ based on the depth map $D$, and $M$ is the warp mask indicating the visible points in both views. Note that the warper $\mathcal{W}_{I\to{J}}$ is a deterministic function when the two views and the depth map are known.

The novel view image $\hat{J}_{\epsilon,t,y_{\varphi_2}}$ is generated from the input view $c_I$ based on the pretrained diffusion model $\epsilon_{pretrain}$ as follows, 
\begin{equation}
\hat{J}_{\epsilon,t,y_{\varphi_2}} = \alpha_tI+\sigma_t\epsilon_{pretrain}(I,t,y_{\varphi_2},D_{J}),
\label{fml5}
\end{equation}
where $\alpha_t$ and $\sigma_t$ are predefined parameters in the pretrained diffusion model $\epsilon_{pretrain}$ conditioned on time step $t$, $D_{J}$ is the estimated depth map under view $c_J$. Here, ControlNet~\cite{zhang_adding_2023} is adopted as the pretrained diffusion model with depth map as conditions. 
With the warp loss, the geometric token enables the diffusion model to have the capability of cross-view generation with the learnable parameter $\varphi_2$. 

In the implementation, we use Point-E~\cite{nichol_point-e_2022} to generate the 3D point cloud and then obtain the depth map of the input image. Initially, we use the input reference view $c_{ref}$ as $c_I$, and then sample a neighboring view as $c_J$ with a small view change. After multiple steps, views from 360 degrees will be sampled.

\begin{figure}[t!]
\centering
\includegraphics[width=\columnwidth]{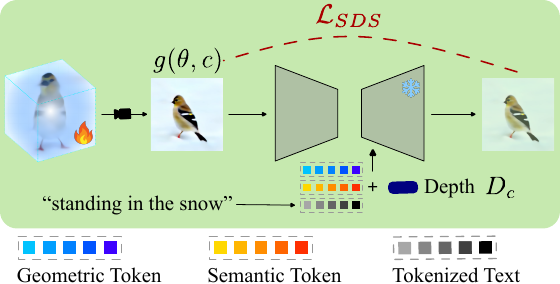}
\caption{\textbf{Score distillation sampling.} A rendered image of a 3D volume is utilized as the input and a depth ControlNet with low CFG scales is utilized for generation. For the text condition, we combine the semantic token and geometric token with tokenized texts, which enables background editing and object editing through prompt.} 
\label{score}
\end{figure}

\subsubsection{Reconstruction Loss}
The reconstruction loss ensure the geometric token $\varphi_2$ to retain subject semantics under the reference view $c_{ref}$ with the reference image $I_{ref}$ as follows, 
\begin{equation}
\begin{split}
\min&_{\varphi_2}\mathcal{L}_{rec}(\varphi_2 \mid c_{ref}):=\mathbb{E}_{x,\epsilon,t}[w(t)\cdot\\
&\lVert\epsilon_{pretrain}(I_{ref}\cdot M_{ref},t,y_{\varphi_2},D_{ref})-\epsilon_t\cdot M_{ref}\lVert_2^2],
\end{split}
\label{fml6}
\end{equation}
where $D_{ref}$ is the depth map image and $M_{ref}$ is the object mask. This enforces the model to generate the ground truth image when guided by the true depth, ensuring consistent subject identity. 


\subsection{Low-scale Score Distillation Sampling}
\label{sec:stage_3}
In the score distillation sampling stage (Fig. \ref{score}), we use prompts $y_{\varphi_1,\varphi_2}$ with both $\varphi_1$ parameterized semantic token and $\varphi_2$ parameterized geometric token, guided by the depth map $D_c$ under the sampled view $c$.
The aim of this stage is to learn a 3D volume parameterized by $\theta$. Specifically, we adopt the deformed SDS formulation as follows:
\begin{equation}
\begin{split}
\nabla_\theta\mathcal{L}_{SDS}&(\theta):= \mathbb{E}_{t,\epsilon,c}[w(t)\cdot\\
&(\epsilon_{pretrain}(x_t,t,y_{\varphi_1,\varphi_2},D_c)-\epsilon_t) \frac{\partial g(\theta,c)}{\partial \theta}],
\end{split}
\label{fml8}
\end{equation}
where the time step $t\sim \mathcal{U}\left(0.02,0.98\right)$, noise $\epsilon_t \sim \mathcal{N}(0,\mathcal{I})$, and $g(\theta,c)$ is the rendered image from the 3D volumes parameterized by $\theta$ under camera view $c$, $x_t = \alpha_t g(\theta,c) + \sigma_t \epsilon$.

The scaling factor $w(t)$ in Eq.{~\ref{fml8}} allows flexibly tuning the degree of conditionality in classifier-free guidance (CFG) for text-to-image generation~\cite{ho_classifier-free_2022}. Higher scales impose stronger conditional constraints, while lower scales enable more unconditional generation. In 2D text-to-image, the CFG scale is typically set between 7.5 and 10 to balance quality and diversity.

Typically, high CFG scales (up to 100) are required for text-to-3D optimization as DreamFusion proposed. However, excessively high scales can impair image quality and diversity~\cite{wang_prolificdreamer_2023}. Our consistency tokens learned in the first two stages enhance semantic and geometric consistency, allowing high-quality distillation even at low scales ($<25$). This achieves photo-realistic, natural-saturated 3D generations faithfully adhering to the subject.

\begin{figure}[t]
\centering
\includegraphics[width=1\linewidth]{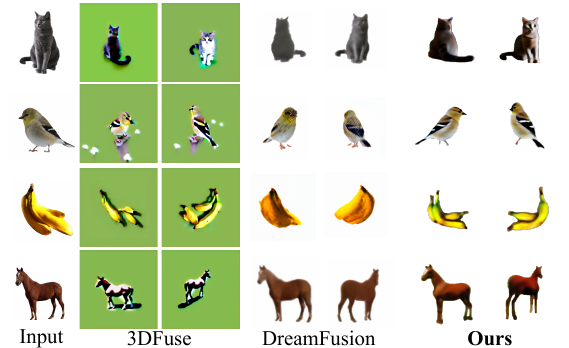}
\caption{\textbf{Comparison with baselines (text-to-3D generation from a single image).} The first column is the single input image. The following columns are results of 3DFuse, DreamFusion and Consist3D(Ours), separately. DreamFusion cannot correctly synthesize photo-realistic images and thin objects. 3DFuse is strongly troubled by inconsistency issues. However, the generation results of our method are not only faithful to the reference but also natural saturated, with good consistencies.}
\label{compare}
\end{figure}
\begin{figure*}[t]
\centering
\includegraphics[width=1\textwidth]{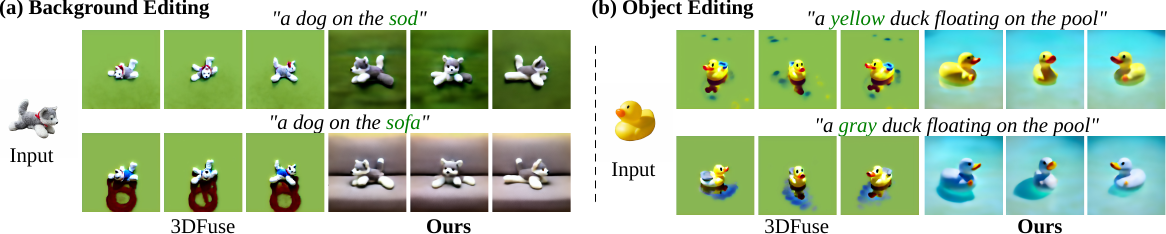} 
\caption{\textbf{Comparison with baselines (background and object editing).} (a) 3DFuse cannot correctly generate the background of the dog, while our method generates ``sod" and ``sofa" properly. (b) With the object ``yellow duck" changed to ``gray duck", 3DFuse only generates a duck with small gray wings, while our method changes the whole body to gray successfully. }
\label{bg_obj}
\end{figure*}
\begin{table*}[h]
\centering
\resizebox{0.95\linewidth}{!}{
\begin{tabular}{l|cccccccccccc|c}
\toprule
Methods & anya & banana & bird & butterfly & cat & clock & duck & hat & horse & shark & sneaker & sunglasses & Average\\
\midrule
DreamFusion & \textbf{80.71} & 69.80 & 71.93 & 65.76 & 72.76 & 69.38 & 80.34 & 71.10 & 75.40 & 67.31 & 63.46 & 60.59 & 70.71\\
3DFuse & 70.31 & 71.72 & 73.41 & 75.12 & 67.19 & 64.60 & 78.84 & 62.63 & 68.52 & 71.81 & 67.33 & 74.05 & 70.46\\
Consist3D (ours) & 80.05 & \textbf{77.22} & \textbf{80.99} & \textbf{82.00} & \textbf{72.85} & \textbf{71.47} & \textbf{84.13} & \textbf{87.34} & \textbf{77.96} & \textbf{72.04} & \textbf{73.59} & \textbf{77.04} & \textbf{78.06}\\
\bottomrule
\end{tabular}
}
\caption{\textbf{CLIP score}. The bold ones in the figure are the highest CLIP Scores in each category. The performance of our method surpasses baselines comprehensively.}
\label{table:clip_score}
\end{table*}
\begin{table}[h]
\centering
\resizebox{\linewidth}{!}{

\begin{tabular}{l|ccccc|c}
\toprule
Method & saturation & geometric & semantic & fidelity & clarity & Average\\
\midrule
DreamFusion & 65.40 & 72.20 & 76.07 & 79.29 & 76.10 & 73.81\\
3DFuse & 69.87 & 73.51 & 63.58 & 72.51 & 77.54 & 71.40\\
Consist3D (ours) & \textbf{89.73} & \textbf{77.29} & \textbf{85.35} & \textbf{82.20} & \textbf{79.36} & \textbf{82.79}\\
\bottomrule
\end{tabular}
}
\caption{\textbf{User study.} The bold ones in the figure are the best scores in each type of problem. Our method surpasses baselines in both consistency and quality.}
\label{table:human}
\end{table}
\section{Experiments}
\subsection{Implementation Details}
We use Stable Diffusion~\cite{rombach_high-resolution_2022} as the generative model with CLIP~\cite{radford_learning_2021} text encoder and LoRA~\cite{hu_lora_2021} as the adapter technique for fine-tuning. As for the representation of the 3D field, we adopt Score Jacobian Chaining (SJC)~\cite{wang_score_2022}. Encoding takes half an hour for each of the two stages, and distillation takes another hour. Specifically, Stage I semantic encoding uses $1k$ optimization steps with LoRA. Stage II geometric encoding uses $2k$ optimization steps with LoRA. Stage III uses $10k$ optimization steps for SJC.

\subsection{Datasets} 
We evaluate Consist3D on a wide range of image collections, where the categories include animals, toys, and cartoon characters, and each subject is presented with a single-view capture. The sources of the image collections includes in-the-wild images selected from ImageNet, cartoon characters collected from the Internet, and images from the DreamBooth3D dataset. We optimize each 3D asset corresponding to the given single-view image with several different prompts and backgrounds, demonstrating faithful and photo-realistic 3D generation results in diversity.

\subsection{Performance Comparison}
\subsubsection{Text-to-3D Generation from a Single Image}
We compare our results with two baselines (\textit{e.g.}, DreamFusion~\cite{poole_dreamfusion_2022} and 3DFuse~\cite{seo_let_2023}) in single-image based text-to-3D generation task, because they are the most related to our method and are representative works in the field of personalized 3D generation. Notably, the original implementation of DreamFusion is a text-to-3D structure. Therefore, we initially utilized a single-view image along with DreamBooth for one-shot tuning, and then incorporated the shape prior estimated by Zero-1-to-3 for DreamFusion to produce 3D assets, following the implementation of the official open source code. 
Our results are bench-marked against the baseline DreamFusion and 3DFuse, as Fig.~\ref{compare} shows. 
In the case of 3DFuse, we adhere to the original configuration established by its authors. We notice that DreamFusion suffers from incorrect shape prior estimations and gives unnatural results (Fig.~\ref{compare}), due to the super-high guidance scale. The generation quality of 3DFuse is strictly limited by the estimated point cloud prior, and even a slightly incorrect depth map can lead to completely wrong generations. Furthermore, its generated objects are not faithful enough to the given reference view, most time with blurred edges and overfitting problems. In contrast, our work can generate objects that are faithful to the input image, and achieve a more natural and photo-realistic reconstruction, not leaning towards the artistic model.

\subsubsection{Background and Object Editing}
For background editing, we compare our results with 3DFuse, since DreamFusion does not provide an option to edit the background. As Fig.~\ref{bg_obj} (a) shows, 3DFuse is unable to correctly generate the correct background, even if we have turned on the background generation option on.  In contrast, our model is capable of editing the background for the reconstructed object by diverse prompts.
For object editing, we compare our results with 3DFuse as well. As Fig.~\ref{bg_obj} (b) shows, 3DFuse is also unable to correctly change the object with the unchanged input image while our model can make it.

\subsection{Quantitative Evaluation}
We compare our method with two baselines under 12 categories as shown in Tab.~\ref{table:clip_score}. The data for these 12 categories include unseen-domain anime from Internet, in-the-wild images from ImageNet, and synthetic datasets from the DreamBooth Dataset. CLIP Score~\cite{hessel_clipscore_2022} is a measure of image-text similarity. The higher the value, the more semantically the image and text match. We extend it to image-to-image similarity, which measures the similarity between the generated result and a given single-view image. To extend and calculate CLIP Score, we first re-describe the reference image with BLIP-2, and for fairness, remove the background description. Then for the reconstruction results of the 3D generation methods, we sample 100 camera positions, and for each position's sampled image, calculate the CLIP Score with the previously obtained description, and take the average separately for each method. Our method far surpasses the baselines in most categories, which means our generations are more faithful to the reference subject.

\subsection{User Study}
We have conducted a user study with 137 participants, which is shown in Table.~\ref{table:human}. We have asked the participants to choose their preferred result in terms of saturation consistency, geometric consistency, semantic consistency, overall fidelity, and overall clarity among DreamFusion, 3DFuse, and Consist3D (Ours). The results show that Consist3D generates 3D scenes that the majority of people judge as having better consistency and overall quality than other methods.

\begin{figure*}[t]
\centering
\includegraphics[width=1\textwidth]{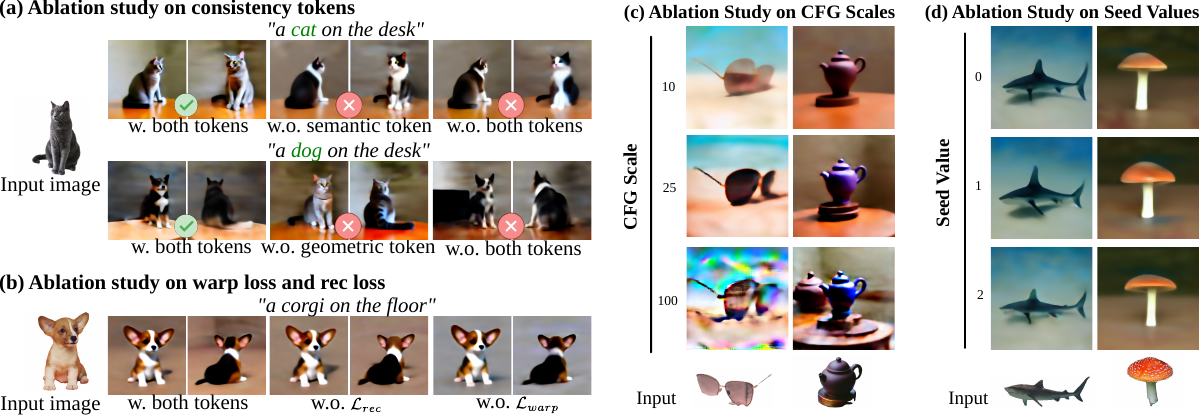} 
\caption{\textbf{Ablation study.} (a) Consistency Tokens. The first row: without semantic token, the generation is not faithful to the input image. The second row: without geometric token, the generation is not consist in novel views and fail to do object editing. (b) Losses. Without $\mathcal{L}_{rec}$ or $\mathcal{L}_{warp}$, the generation becomes not faithful to the input image or fails to generate correct background. (c) CFG Scales. The 3D synthesis of sunglasses and teapot under CFG scales of 10, 25, 100, separately. Results at lower scale are more natural saturated, while the generations at higher scale tend to be with over-saturated color. (d) Seed Values. The 3D synthesis of shark and mushroom under seed values of 0, 1, 2 demonstrates our method is robust to different seeds, with the generations slightly changed.} 
\label{ablation}
\end{figure*}
\subsection{Ablation Studies} 

\subsubsection{Consistency Tokens}
In Fig.~\ref{ablation} (a), we show ablation studies for the two consistency tokens. First, we test the role of semantic token on single-image text-to-3D generation, and we find that removing the semantic token will cause the generated object's semantics to be inconsistent with the input image. 
In addition, we test the role of geometric token on object editing. We find that removing the geometric token leads to inconsistent generations under novel views, and the generated object could not be edited, which indirectly proves that the geometric token emphasizes the shape's geometric consistency across different viewing angles, while not overfitting to the view of the input image.

\subsubsection{Losses}
In Fig.~\ref{ablation} (b), we test the roles of warp loss $\mathcal{L}_{warp}$ and reconstruction loss $\mathcal{L}_{rec}$ in the geometric encoding stage. With only the loss $\mathcal{L}_{rec}$ applied, the model cannot generate correct background but the object is faithful to the input image. With only the loss $\mathcal{L}_{warp}$ applied, the model generates geometric consistent 3D asset, but the background is not faithful to the input text. With both $\mathcal{L}_{warp}$ and $\mathcal{L}_{rec}$ applied, the results are faithful to the input image and text prompt while keeping geometric consistency with object and background correctly generated.

\subsubsection{CFG Scales}
In the ablation study for score distillation sampling (Fig.~\ref{ablation} (c)), we vary the CFG scale to 10, 25, and 100, showing our low-scale distillation improves 3D quality. The consistency tokens reduce scale requirements for photo-realistic 3D generation. Our method achieves good, diverse results with low scales (below 25).

\subsubsection{Seed Values}
We experiment with fixed prompts and changing random seeds to verify the robustness of our approach as Fig.~\ref{ablation} (d) shows. The results demonstrate that our approach is not sensitive to random seeds.

\section{Limitation and Future Work}
Our method fails when the point cloud estimation is severely distorted. Moreover, if overly complex background prompts are used, the model may not be able to generate high-detail backgrounds. 
In future work, we intend to model objects and backgrounds separately to obtain more refined generations.

\section{Conclusion}
We introduce Consist3D, a method to faithfully and photo-realistically personalize text-to-3D generation from a single view image, with background and object editable by text prompts.  addressing the inconsistency issues in semantic, geometry, and saturation.
Specifically, we propose a 3-stage framework with a semantic encoding, a geometric encoding stages and a low-scale score distillation sampling stage. 
The semantic token learned in the first encoding stage encourages Consist3D to be robust to shape estimation, the geometric token learned in the second stage encourages the generation to be consist across different views, and both of the token are used in the third stage to encourage natural saturation of the 3D generation.
Our method outperforms baselines quantitatively and qualitatively. Experiments on a wide range of images (including in-the-wild and synthesis images) demonstrate our approach can 1) generate high-fidelity 3D assets with consistency from one image. 2) change the background or Object of the 3D generations through editing the text prompts without changing the input image.
Going forward, we plan to incorporate more geometric constraints into token training to further enhance 3D consistency.

\bibliography{aaai24}

\end{document}